\documentclass{article}
\usepackage{spconf,amsmath,graphicx}
\usepackage[dvipsnames]{xcolor}
\usepackage{siunitx}
\usepackage{algorithm}
\usepackage{algpseudocode}
\usepackage{subcaption}
\usepackage{enumitem}

\newcommand\blfootnote[1]{%
  \begingroup
  \renewcommand\thefootnote{}\footnote{#1}%
  \addtocounter{footnote}{-1}%
  \endgroup
}

\title{CAN WE AVOID DOUBLE DESCENT IN DEEP NEURAL NETWORKS?}
\name{Victor Qu\'etu \qquad Enzo Tartaglione}
\address{LTCI, T\'el\'ecom Paris, Institut Polytechnique de Paris}

\begin{document}

\maketitle

\begin{abstract}

Finding the optimal size of deep learning models is very actual and of broad impact, especially in energy-saving schemes. Very recently, an unexpected phenomenon, the ``double descent'', has caught the attention of the deep learning community. As the model's size grows, the performance gets first worse and then goes back to improving. It raises serious questions about the optimal model's size to maintain high generalization: the model needs to be sufficiently over-parametrized, but adding too many parameters wastes training resources. Is it possible to find, in an efficient way, the best trade-off?\\
Our work shows that the double descent phenomenon is potentially avoidable with proper conditioning of the learning problem, but a final answer is yet to be found. We empirically observe that there is hope to dodge the double descent in complex scenarios with proper regularization, as a simple $\ell_2$ regularization is already positively contributing to such a perspective.
\end{abstract}

%\blfootnote{© 2023 IEEE. Personal use of this material is permitted. Permission from IEEE must be obtained for all other uses, in any current or future media, including reprinting/republishing this material for advertising or promotional purposes, creating new collective works, for resale or redistribution to servers or lists, or reuse of any copyrighted component of this work in other works.}
%\blfootnote{Article accepted for publication by the International Conference on Image Processing (ICIP23).}
\blfootnote{© 2023 IEEE. Personal use of this material is permitted. Permission from IEEE must be obtained for all other uses, in any current or future media, including reprinting/republishing this material for advertising or promotional purposes, creating new collective works, for resale or redistribution to servers or lists, or reuse of any copyrighted component of this work in other works. \\

\noindent Article accepted for publication at the International Conference on Image Processing (ICIP23).}
\section{The race towards generalization}
\label{sec:Introduction}

In the recent few years, the research community has witnessed a race towards achieving higher and higher performance (in terms of an error on unseen data), proposing very large architectures like, for example, transformers~\cite{liu2021swin}. From big architectures come big responsibilities: learning strategies to avoid the over-fitting urge to be developed.\\
The most straightforward approach would be to provide more data: deep learning methods are notoriously data hunger. Since they typically optimize some objective function through gradient descent, having more data in the training set helps the optimization process in selecting the most appropriate set of features (to oversimplify, the most recurrent ones). This allows us to have high performance on unseen data. Such an approach has the big drawbacks of requiring enormous computational power for training and, most importantly, large annotated datasets. While undertaking the first drawback is a hot research topic~\cite{frankle2018the, bragagnolo2022update, tartaglione2022rise}, the second is broadly explored with approaches like transfer learning~\cite{zhuang2020comprehensive} or self-supervised learning~\cite{ravanelli2020multi}.
\begin{figure}[t]
    \includegraphics[width=\linewidth]{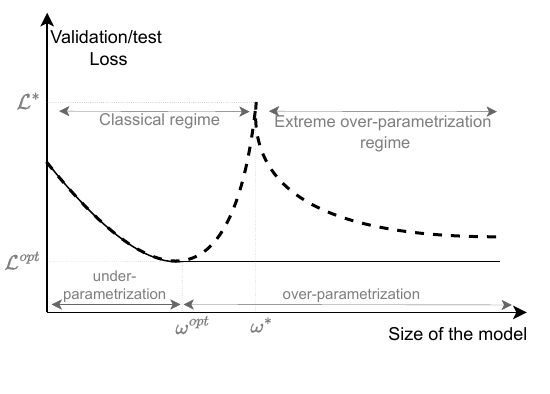}
    \caption{The double descent phenomenon (dashed line): is it possible to constrain the learning problem to a minimum such that the loss, in over-parametrized regimes, remains close to $\mathcal{L}^{opt}$ (continuous line)?}
    \label{fig:DD_scheme}
\end{figure}

\noindent In more realistic cases, large datasets are typically unavailable, and in those cases approaches working with small data, in the context of \emph{frugal AI}, need to be employed. This poses research questions on how to enlarge the available datasets or to transfer knowledge from similar tasks; however, it also poses questions on how to optimally dimension the deep learning model to be trained. Contrarily to what is expected from the bias-variance trade-off, the phenomenon of \emph{double descent} can be observed in a very over-parameterized network: given some optimal set of parameters for the model $\boldsymbol{w}^{opt}$ with loss value $\mathcal{L}^{opt}$, adding more parameters will worsen the performance until a local maximum $\mathcal{L}^{*}$ beyond which, adding even more parameters, the performance is back improved. This phenomenon, named \emph{double descent}~\cite{Belkin_2019} is displayed in Fig.~\ref{fig:DD_scheme}, and is consistently reported in the literature~\cite{spigler2019jamming, Geiger_2019}. Double descent poses the serious quest of finding the best set of parameters, in order not to fall into an over-parametrized (or under-parametrized) regime. The possible approaches to tackle this are two: finding $\boldsymbol{w}^{opt}$, which is a problem typically requiring a lot of computation, or extremely over-parametrizing the model. Unfortunately, both the roads are not compatible with a frugal setup: is there a solution to this problem?\\
In this work, we show that the double descent phenomenon is potentially avoidable. Having a sufficiently large regularization on the model's parameters drives the deep models in a configuration where the set of parameters in excess $\boldsymbol{w}^{exc}$ are essentially producing no perturbation on the output of the model. Nevertheless, as opposed to  Nakkiran~et~al.~\cite{nakkiran2021optimal}, who showed in regression tasks that such a regularization could help in dodging double descent, we observe that, in classification tasks, this regularization is typically insufficient in standard scenarios: an ingredient is still missing, although we are on the right path.

\section{Double descent and its implications}
\label{sec:Related work}

\textbf{Double descent in machine learning models.} The double descent phenomenon has been highlighted in various machine learning models, like decision trees, random features~\cite{Meng-Random-Feature-Model}, linear regression~\cite{muthukumar2020harmless} and deep neural networks~\cite{yilmaz2022regularization}. Based on the calculation of the precise limit of the excess risk under the high dimensional framework where the training sample size, the dimension of data, and the dimension of random features tend to infinity proportionally, Meng~et~al.~\cite{Meng-Random-Feature-Model} demonstrate that the risk curves of double random feature models can exhibit double and even multiple descents. The double descent risk curve was proposed to qualitatively describe the out-of-sample prediction accuracy of variably-parameterized machine learning models. Muthukumar~et~al.~\cite{muthukumar2020harmless} provide a precise mathematical analysis for the shape of this curve in two simple data models with the least squares/least norm predictor. Nakkiran~et~al.~\cite{nakkiran2021deep} showed that the double descent phenomenon is not limited to varying the model size, but it is also observed as a function of training time or epochs and also identified certain regimes were increasing the number of train samples hurts test performance.\\
\textbf{Double descent in regression tasks.} It has been recently shown that, for certain linear regression models with isotropic data distribution, optimally-tuned $\ell_2$ regularization can achieve monotonic test performance as either the sample size or the model size is grown. Nakkiran~et~al.~\cite{nakkiran2021optimal} demonstrated it analytically and established that optimally-tuned $\ell_2$ regularization can mitigate double descent for general models, including neural networks like Convolutional Neural Networks. Endorsing such a result, Yilmaz~et~al.~\cite{yilmaz2022regularization} indicated that regularization-wise double descent can be explained as a superposition of bias-variance trade-offs on different features of the data (for a linear model) or parts of the neural network and that double descent can be eliminated by scaling the regularization strengths accordingly.\\
\textbf{Double descent for classification tasks.}  Although many signs of progress in regression models have been done, in classification tasks, the problem of avoiding, or formally characterizing, the double descent phenomenon, is much harder to tackle. The test error of standard deep networks, like the ResNet architecture, trained on standard image classification datasets, consistently follows a double descent curve both when there is label noise (CIFAR-10) and without any label noise (CIFAR-100)~\cite{yilmaz2022regularization}. Double descent of pruned models concerning the number of original model parameters has been studied by Chang~et~al, which reveals a double descent behavior also in model pruning~\cite{Chang_Overparameterization}. Model-wise, the double descent phenomenon has been studied a lot under the spectrum of over-parametrization: a recent work also confirmed that sparsification via network pruning can cause double descent in the presence of noisy labels~\cite{SparseDoubleDescent}. He~et~al.~\cite{SparseDoubleDescent} proposed a novel learning distance interpretation, where they observe a correlation between the model's configurations before and after training with the sparse double descent curve, emphasizing the flatness of the minimum reached after optimization. Our work differs from this study by enlightening some cases in which the double descent phenomenon is not evident. We show, in particular, that by imposing some constraints on the learning process, we can avoid the double descent. The experimental setup follows the same as He~et~al.'s.

\section{Dodging the double descent}
\label{sec:method}

\begin{figure*}[t]
    \centering
    \begin{subfigure}{.685\columnwidth}
        \includegraphics[width=\columnwidth]{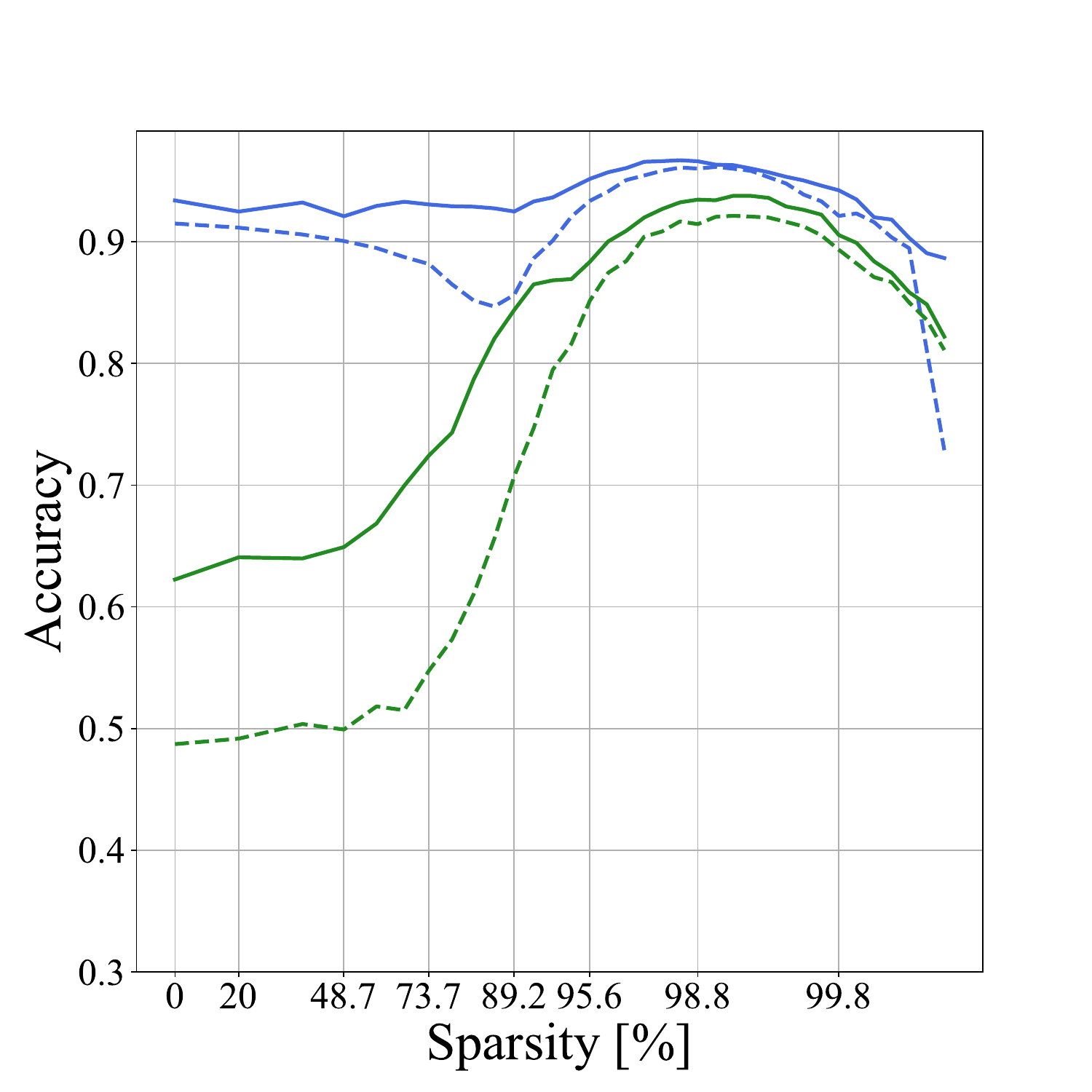}
    \end{subfigure}\hfill
    \begin{subfigure}{.685\columnwidth}
        \includegraphics[width=\columnwidth]{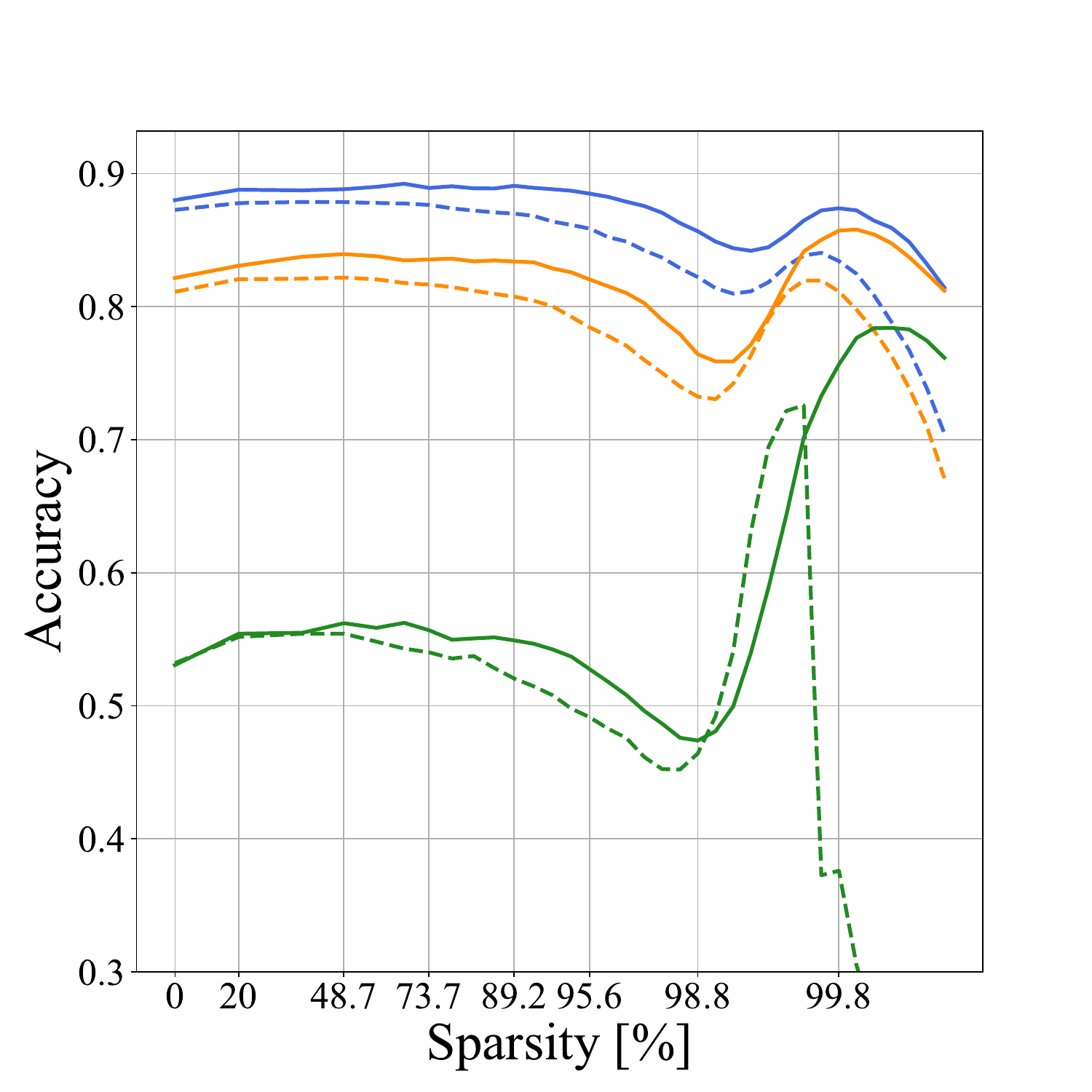}
    \end{subfigure}
    \begin{subfigure}{.685\columnwidth}
        \includegraphics[width=\columnwidth]{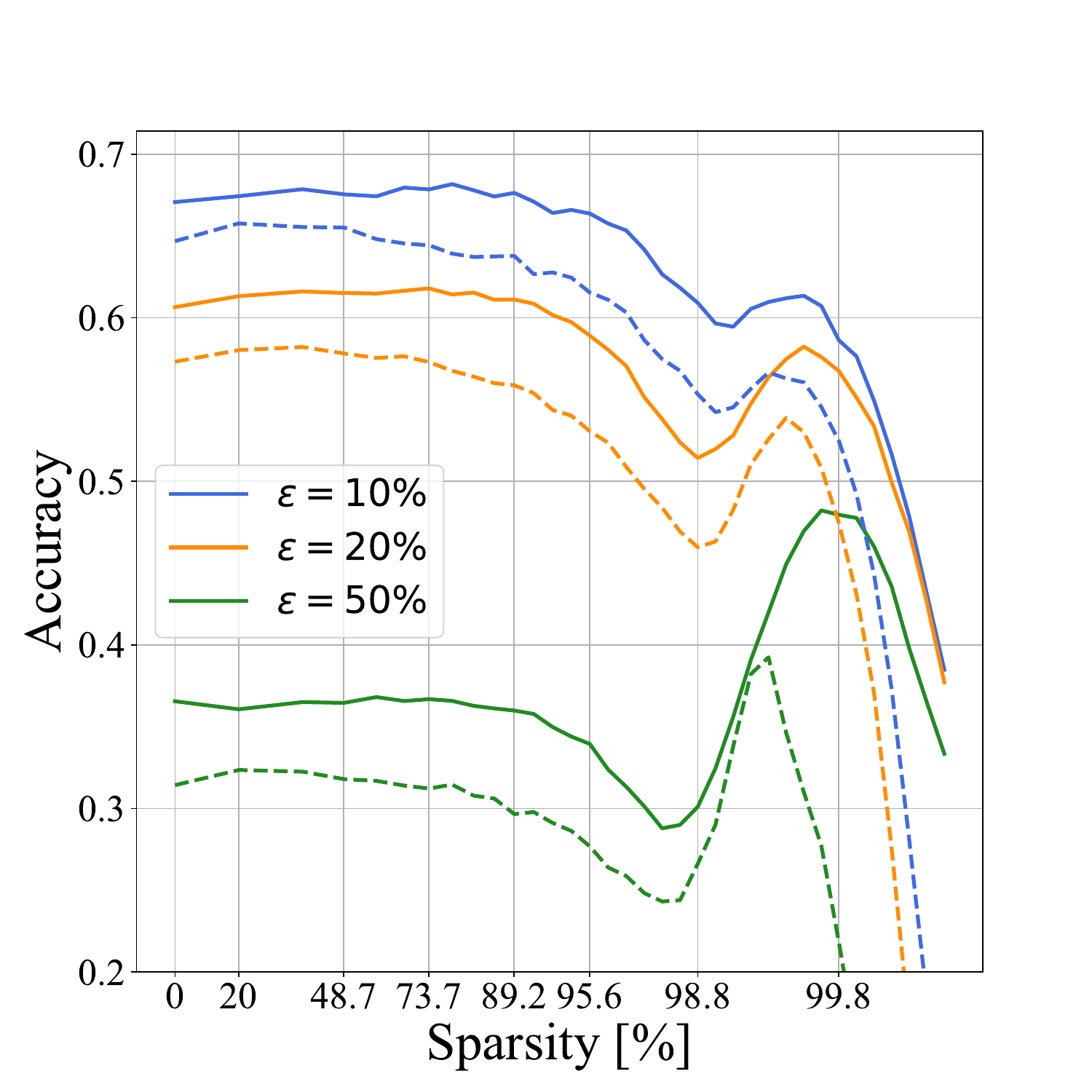}
    \end{subfigure}
    \caption{Test accuracy in the function of sparsity with different symmetric noise $\varepsilon$. Dashed lines are vanilla and solids lines to $\ell_2$ regularization. 
    \textbf{Left:} LeNet-300-100 on MNIST.
    \textbf{Middle:} ResNet-18 on CIFAR-10.
    \textbf{Right:} ResNet-18 on CIFAR-100.}
    \label{fig:MNIST}
\end{figure*}
\begin{algorithm}[t]
    \caption{Drawing performance in over-parametrized \\setups: \textbf{P}rune, \textbf{R}ewind, \textbf{T}rain, repeat.}
    \label{Algo}
    \begin{algorithmic}[1]
        \Procedure{\textbf{PRT}($\boldsymbol{w}^{init}$, $\Xi$, $\lambda$, $T^{iter}$,$T^{end}$)}{}
            \State $\boldsymbol{w} \gets$ Train($\boldsymbol{w}^{init}$, $\Xi$, $\lambda$)\label{line:dense}
            \While{Sparsity($\boldsymbol{w}, \boldsymbol{w}^{init}$) $< T^{end}$}\label{line:endcond}
                \State $\boldsymbol{w} \gets$ Prune($\boldsymbol{w}$, $T^{iter}$) \label{line:prune}
                \State $\boldsymbol{w} \gets$ Rewind($\boldsymbol{w}$, $\boldsymbol{w}^{init}$)\label{line:rewind}
                \State $\boldsymbol{w} \gets$ Train($\boldsymbol{w}$,$\Xi$, $\lambda$)\label{line:wd} 
            \EndWhile
        \EndProcedure
    \end{algorithmic}
\end{algorithm}
\textbf{A regularization-based approach.} In the previous section we have presented all the main contributions achieved in the literature around the double descent phenomenon. It is a known result that, given an over-parametrized model, without special constraints on the training, this phenomenon can be consistently observed. However, it is also known that, for an optimally parametrized model, double descent should not occur. Let us say, the optimal output for the model (what we aim at), is $\boldsymbol{y}^{opt}$. For instance, there is some subset $\boldsymbol{w}^{exc}\in \boldsymbol{w}$ of parameters belonging to the model which are in excess, namely the ones contributing to the double descent phenomenon. Since these are not essential in the learning/inference steps, they can be considered as \emph{noisy parameters}, which deteriorate the performance of the whole model and make the whole learning problem more difficult. They generate a perturbation in the output of the model which we can quantify as
\begin{equation}
    \boldsymbol{y}^{exc} = \sum_{w_i\in \boldsymbol{w}^{exc}} \text{forward}\left[\phi(\boldsymbol{x}_i\cdot w_i)\right],
\end{equation}
where $\boldsymbol{x}_i$ is the input(s) processed by $w_i$, $\phi(\cdot)$ is some non-linear activation for which $\phi(z)\approx z$ when $z\rightarrow 0$, and $\text{forward}(\cdot)$ simply forward-propagates the signal to the output of the neural network. As the output of the model is given by $\boldsymbol{y}=\boldsymbol{y}^{opt}+\boldsymbol{y}^{exc}$, in the optimal case we would require that $\boldsymbol{y}^{exc}=\boldsymbol{0}$. To satisfy such a condition, we have two possible scenarios:
\begin{itemize}[noitemsep,topsep=0pt]
    \item $\exists w_i\neq 0 \in \boldsymbol{w}^{exc}$. In this case, the optimizer finds a minimum loss such that the algebraic sum of the noisy contribution is zero. When a subset of these is removed, it is possible that $\boldsymbol{y}^{exc}\neq \boldsymbol{0}$, which results in a performance loss and, for instance, the double descent phenomenon is observed. 
    \item $w_i=0, \forall w_i\in \boldsymbol{w}^{exc}$. In this case, there is no contribution of the noisy terms and we are in the optimal scenario, where the parameters are de-facto removed from the model. 
\end{itemize}
Let us focus on the second case. For numerical reasons, this scenario is unrealistic during optimization; hence we can achieve a similar outcome by satisfying two conditions:
\begin{enumerate}[noitemsep,topsep=0pt]
    \item $w_i x_i \approx 0, \forall w_i \in \boldsymbol{w}^{exc}$;
    \item $\|\text{forward}\left[\phi(\boldsymbol{x}_i\cdot w_i)\right]\|_1 \leq \|\phi(\boldsymbol{x}_i\cdot w_i)\|_1$,\\ with $\|\phi(\boldsymbol{x}_i\cdot w_i)\|_1\approx 0$. 
\end{enumerate}
We can achieve these conditions with a sufficiently large regularization on the parameters $\boldsymbol{w}$. The first condition is achievable employing any typical weight penalty as we assume that, for local minima of the loss where $\frac{\partial L}{\partial w_i} = 0$, we have some weight penalty $C$ pushing the parameter's magnitude towards zero. The second condition, on the contrary, requires more careful consideration. Indeed, for the propriety of the function's composability, we need to ensure that the activation function in every layer does not amplify the signal (true in the most common scenarios) and that all the parameters have the lowest magnitude possible. For this reason, ideally, the regulation to employ should be $\ell_\infty$. On the other hand, however, we are also required to enforce some sparsity: towards this end, recent works in the field suggest $\ell_2$ regulation is a fair compromise~\cite{han2015learning}. Hence, we just need a sufficiently large $\ell_2$ regularization.\\
\textbf{How can we observe we have avoided double descent?} We present an algorithm to draw the (eventual) double descent phenomenon in Alg.~\ref{Algo}. After training for the first time the model on the learning task $\Xi$, eventually with $\ell_2$ regularization scaled by $\lambda$ (line~\ref{line:dense}), a magnitude pruning stage is set up (line~\ref{line:prune}). Neural network pruning, whose goal is to reduce a large network to a smaller one without altering accuracy, removes irrelevant weights, filters, or other structures from neural networks. An unstructured pruning method called magnitude-based pruning, popularized by~\cite{han2015learning}, adopted a process in which weights, below some specific threshold $T$, are pruned (line~\ref{line:prune}). We highlight that more complex pruning approaches exist, but magnitude-based pruning shows its competitiveness despite very low complexity~\cite{Gale_Magnitude}. Towards this end, the hyper-parameter $T^{iter}$ sets the relative pruning percentage, or in other words, how many parameters will be removed at every pruning stage. 
Once pruned, the accuracy of the model typically decreases. To recover performance, the lottery ticket rewinding technique, proposed by Frankle \& Carbin~\cite{LTR}, is employed. It consists of retraining the subset of parameters that are surviving the pruning stage to their initialization value (line~\ref{line:rewind}) and then training the model (line~\ref{line:wd}). This approach allows us to state whether a lowly-parametrized model from initialization can, in the best case, learn a target task. We end our drawing once we reach a sparsity higher or equal than $T^{end}$ (line~\ref{line:endcond}).\\ 
\textbf{Experimental setup.} For the experimental setup, we follow the same approach as He~et~al.~\cite{SparseDoubleDescent}. The first model we train is a LeNet-300 on MNIST, for 200 epochs, optimized with SGD with a fixed learning rate of 0.1. The second model is a ResNet-18, trained on CIFAR-10 and on CIFAR-100, for 160 epochs, optimized with SGD, having momentum 0.9 and a learning rate of 0.1, decayed by a factor 0.1 at milestones 80 and 120. For each dataset, a percentage $\varepsilon$ of symmetric, noisy labels is introduced: the labels of a given proportion of training samples are flipped to one of the other class labels, selected with equal probability~\cite{Noisy_labels}. In our experiments, we test with $\varepsilon \in \{10\%, 20\%, 50\%\}$.
\begin{figure}[t]
    \centering
    \begin{subfigure}{1\columnwidth}
        \includegraphics[width=1.0\linewidth]{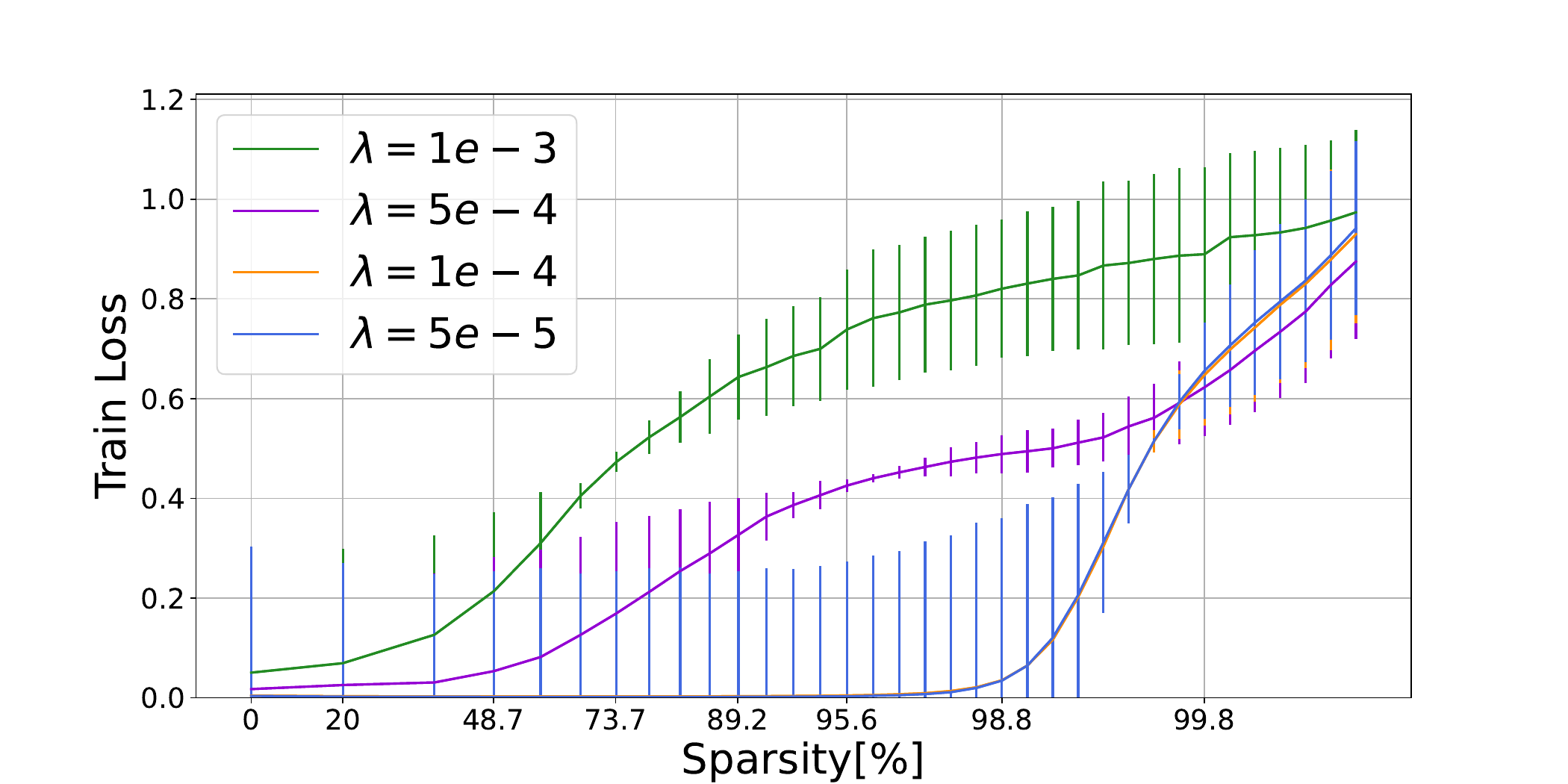}
    \end{subfigure}
    \begin{subfigure}{1\columnwidth}
        \includegraphics[width=1.0\linewidth]{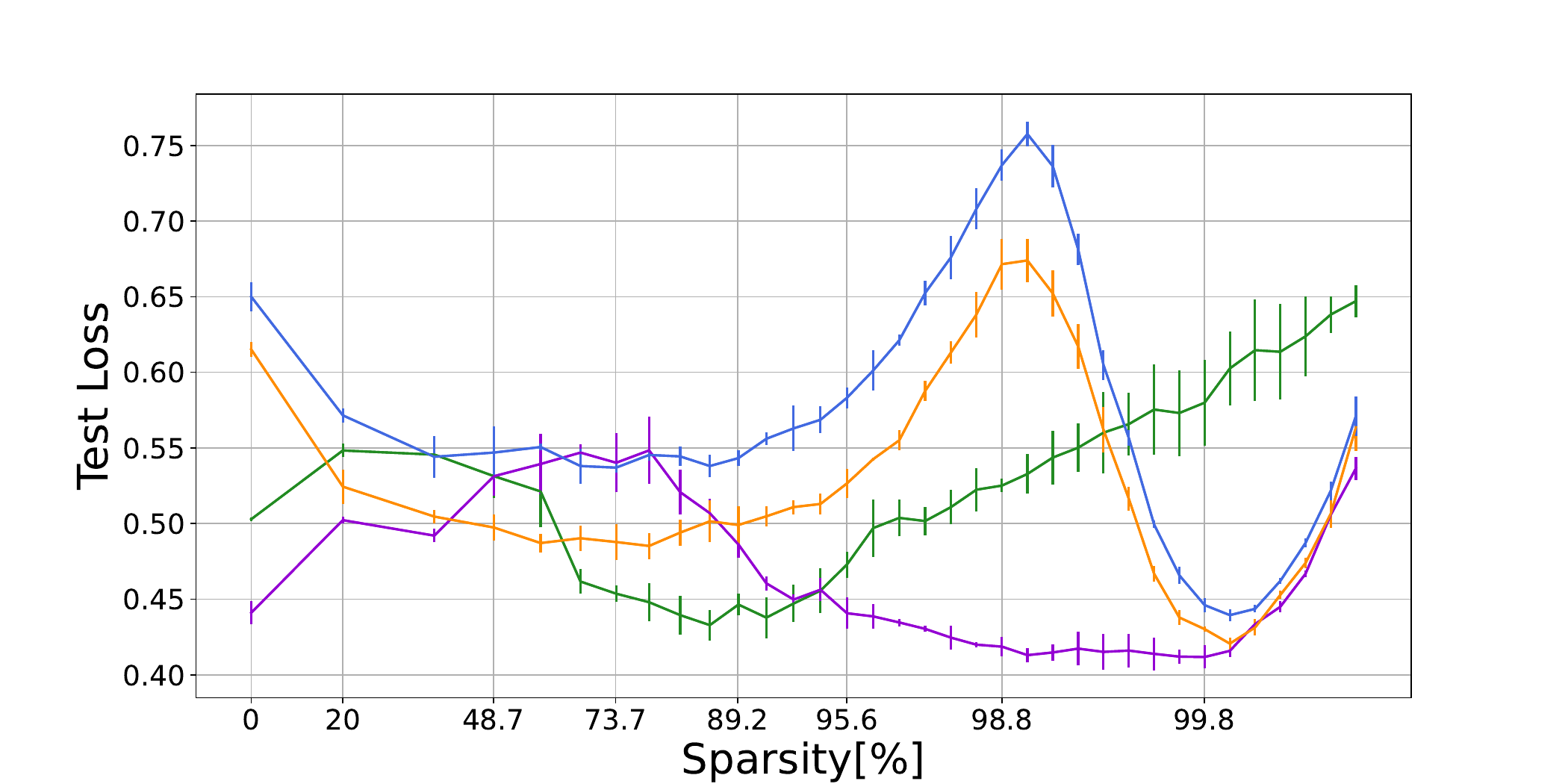}
    \end{subfigure}
    \caption{Train and test loss at varying $\lambda$ for CIFAR-10, with $\varepsilon=10\%$.}
    \label{fig:ablation_test}
\end{figure}
When the regularization $\ell_2$ is employed, we set $\lambda$ to $10^{-4}$. In all the experiments, we use batch sizes of 128 samples, $T^{iter}$ to 20\% and $T^{end}$ is 99.9\%.\\
\textbf{Results.} Fig.~\ref{fig:MNIST} displays our results. As in He~et~al.'s work~\cite{SparseDoubleDescent}, looking at LeNet-300 with $\varepsilon=10\%$, the double descent consists of four phases. First, at low sparsities, the network is massively over-parameterized-thus, the pruned network has a performance similar to the dense one. The second phase is near the interpolation threshold, where the training accuracy is going to drop, and test accuracy is about to first decrease and then increase as the sparsity grows. The third phase is located at high sparsity, where the test accuracy is rising. The final phase happens when both training and test accuracy drop significantly. However, while we can observe the double descent in the test accuracy without $\ell_2$ regularization, the phenomenon fades when the regularization is added. Indeed, the existence of the second phase is questioned: the test accuracy, expected to decrease in this phase, reaches a plateau before rising when regularization is added. In such a simple setup, the double descent is here dodged. However, Fig.~\ref{fig:MNIST} also portrays the result of ResNet-18 on CIFAR-10 and CIFAR-100, with different percentages of noisy labels. Whether the regularization is used or not, on average, and for every value of $\varepsilon$, the double descent phenomenon still occurs. These experiments, which can be considered more complex than the previous one, highlight some limits of the use of standard regularization to avoid the double descent phenomenon and suggest that a specific learning strategy should be designed.\\ 
\textbf{Ablation on $\boldsymbol{\lambda}$.}
In Fig.~\ref{fig:MNIST} we have proposed solutions with a $\ell_2$ regularization hyper-parameter which provides a good trade-off between performance in terms of validation error and avoidance of the double descent.
However, for more complex datasets and more complex convolutional neural networks, such a trade-off can not be easily found: we propose in Fig.~\ref{fig:ablation_test} an ablation study on $\lambda$ for ResNet-18 trained on CIFAR-10 with $\varepsilon=10\%$.
We observe that even for extremely high values of $\lambda$, the double descent is not entirely dodged, and also the performance of the model drops: the imposed regularization becomes too high, and the training set is not entirely learned anymore. This indicates that, while for regression tasks $\ell_2$ regularization is the key to dodging the double descent, in classification tasks, the learning scenario can be very different, and some extra ingredient is yet to be found.

\section{Is double descent avoidable?}
\label{sec:Conclusion}

The problem of finding the best-fitting set of parameters for deep neural networks, which has evident implications for both theoretical and applicative cases, is currently a subject of great interest in the research community. In particular, the phenomenon of double descent prioritizes the research around finding the optimal size for deep neural networks: if a model is not extremely over-parametrized, it may fall into a sub-optimal local minimum, harming the generalization performance.\\
In this paper, we have moved some first steps, in a traditional classification setup, towards avoiding the double descent. If we successfully achieve a local minimum where, regardless of its over-parametrization, the performance of the model is consistently high at varying the cardinality of its parameters, there would not be a strict need in finding the optimal subset of parameters for the trained model. A standard regularization approach like $\ell_2$, which has proven its effectiveness in regression tasks, evidenced some limits in more complex scenarios although dodging effectively the double descent in simpler setups. This result gives us hope: a learning strategy to avoid the double descent might be designed, and will be the subject of future research.

\section{Acknowledgments} 
This project was provided with computer and storage resources by GENCI at IDRIS thanks to the grant 2022-AD011013930 on the supercomputer Jean Zay's the V100 partition.

\bibliographystyle{IEEEbib}
\bibliography{main}

\end{document}